\documentclass[runningheads]{llncs}
\usepackage{booktabs}						% professional-quality tables
\usepackage{multirow}						% tabular cells spanning multiple rows
\usepackage{amsfonts}						% blackboard math symbols
\usepackage{graphicx}						% figures
\usepackage{duckuments}						% sample images        
\usepackage[labelfont=bf]{caption}
\usepackage{subcaption}
\usepackage{dsfont}
\usepackage{multirow}

\begin{document}
\title{Evaluating Neighbor Explainability for Graph Neural Networks}
\author{Oscar Llorente \inst{1} \and
Rana Fawzy\inst{1} \and
Jared Keown\inst{2} \and
Michal Horemuz\inst{2} \and
Péter Vaderna\inst{3} \and
Sándor Laki\inst{4} \and
Roland Kotroczó\inst{4} \and
Rita Csoma\inst{4} \and
János Márk Szalai-Gindl
}
\authorrunning{O. Llorente et al.}
\institute{Ericsson Cognitive Network Solutions, Madrid-Cairo, Spain-Egypt \\ \email{\{oscar.llorente.gonzalez,rana.fawzy\}@ericsson.com} \and
Ericsson Global Artificial Intelligence Accelerator, Stockholm, Sweden \\ \email{\{jared.keown,michal.horemuz\}@ericsson.com} \and
Ericsson Research, Stockholm, Sweden \\ \email{Peter.Vaderna@ericsson.com} \and
Etvös Loránd University, Budapest, Hungary \\ \email{\{lakis,kotroczo.roland,gq92l5,szalaigindl\}@inf.elte.hu}
}
\maketitle  

\begin{abstract}
    Graph Neural Networks (GNNs) have rapidly emerged as powerful tools for modeling complex data structures, particularly in the context of telecommunications, chemistry and social networking. Explainability in GNNs holds essential significance as it empowers stakeholders to gain insights into the inner workings of these complex models, fostering trust and transparency in decision-making processes. In this publication, we address the problem of determining how important is each neighbor for the GNN when classifying a node and how to measure the performance for this specific task. To do this, various known explainability methods are reformulated to calculate the neighbor importance and four new metrics, that aid in determining an explainability method's reliability, are presented. Our results show that there is almost no difference between the explanations provided by gradient-based techniques in the GNN domain, in contrast to other areas of deep learning where this is an active area of research. This means that efforts in this direction may not produce such promising results for GNNs. In addition, many explainability techniques failed to identify important neighbors when GNNs without self-loops are used\footnote{The code to replicate our findings will be available here: \url{https://github.com/EricssonResearch/gnn-neighbors-xai}}.
\end{abstract}

\section{Introduction}
\label{sec: introduction}

In the last decade the area of deep learning has revolutionized the field of machine learning and artificial intelligence, having world-wide impact in applications such as computer vision~\cite{kirillovSegmentAnything2023}, natural language processing~\cite{brownLanguageModelsAre2020c}  or protein folding~\cite{jumperHighlyAccurateProtein2021}. Moreover, in fields dominated by graph-data, as chemistry~\cite{reiserGraphNeuralNetworks2022}, telecommunications networks~\cite{shenGraphNeuralNetworks2021} or social networks~\cite{fanGraphNeuralNetworks2019},  Graph Neural Networks (GNNs) have gained increased popularity, showing great performance and becoming the de facto standard.

However, the main drawback of deep learning methods is the lack of interpretability and explainability. As state of the art models continue to grow in number of parameters and computational complexity, it is increasingly difficult to understand what these techniques are learning and what parts of the data are being used to classify or make a prediction. This is especially sensitive in fields such as autonomous driving~\cite{dikmenTrustAutonomousVehicles2017} or medicine~\cite{amannExplainabilityArtificialIntelligence2020}, where the performance of these techniques could affect human lives directly. To solve this issue, the field of explainability has been growing in recent   years. Here, we follow the distinction presented in \cite{molnarInterpretableMachineLearning}, 
where interpretability and explainability are treated as two different concepts. Interpretability involves constructing a model that is understandable by humans. For deep learning this consists of developing layers, such as the attention layer used in \cite{vaswaniAttentionAllYou2017a}, that allow understanding what is the model focusing on to make its predictions. On the other hand, explainability tries to explain the model regardless of how it is constructed, generating \textit{post hoc} explanations. Several techniques have been presented with that purpose, as in  \cite{simonyanDeepConvolutionalNetworks2014a}, \cite{smilkovSmoothGradRemovingNoise2017}, \cite{zhouLearningDeepFeatures2016} or \cite{selvarajuGradCAMVisualExplanations2017}. Furthermore, there have been improvements in explainability for GNNs in the last years too. For interpretability the mechanism of attention has been adapted to GNNs in \cite{velickovicGraphAttentionNetworks2018}, \cite{brodyHowAttentiveAre2021} or \cite{shiMaskedLabelPrediction2021}. Meanwhile, for explainability some techniques for other domains have been adapted in \cite{baldassarre2019explainability} or \cite{pope2019explainability}, and new techniques have emerged as \cite{ying2019gnnexplainer}, \cite{luo2020parameterized} or \cite{vu2020pgm}.

Recently, a new trend that is starting to appear is building metrics and frameworks to test explainability methods. In the first techniques that were proposed in \cite{simonyanDeepConvolutionalNetworks2014a} or \cite{smilkovSmoothGradRemovingNoise2017} the only evaluation that was done was merely visual. That causes several problems, as pointed out in \cite{adebayoSanityChecksSaliency2018a}. The first one is that only visual and text data can be evaluated that way since it is impossible to test explainability in graphs following that scheme. The other problem is that by evaluating the techniques based on human interpretation of them, a bias could be introduced. It could be the case that a neural network is using different types of features than what a human would use for a given task. In such scenarios, a useful explainability technique should be able to highlight the attributes of the neural network’s decision rather than the preconceived notions of their human users. Some methods have been developed to test techniques as in \cite{hookerBenchmarkInterpretabilityMethods2019a}. While \cite{yuan2022explainability} have also tried to evaluate the performance of the available explainability techniques for GNNs, the difference made by our study is highlighted in Section~\ref{sec: loyalty}.

GNNs can be used to solve several problems, including graph classification, node classification or link prediction. Given that the nature of model interpretation varies significantly depending on the specific problem and application domain, each task involving GNNs necessitates its own tailored set of explainability methods. In node classification, the following approaches can be found in the literature:
\begin{itemize}
    \item Determine the most important features for the GNN, as it is tackled in \cite{ying2019gnnexplainer} or \cite{luo2020parameterized}.
    \item Construct a sub-graph, i.e., a sub-set of important nodes, addressed in \cite{ying2019gnnexplainer} or \cite{luo2020parameterized} too. This can be used to reduce the computation while preserving the most important information for the model (neighbors that it is using most heavily for the classification). However, if the objective is to explain the classification of a single node, defining a sub-graph does not provide detailed information such as relative importance ranking of neighbors, that can be used to understand how the model is behaving.
    \item \cite{baldassarre2019explainability} presented an example of determining the importance of each neighbor and each edge for the classification of a node performed by a GNN. 
\end{itemize}

It is important to clarify here, that following the trend of the literature, as in \cite{ying2019gnnexplainer} or  \cite{luo2020parameterized}, we are using the word importance referring how much a feature or a neighbor affects the classification of a certain node by the GNN. Therefore, this importance is linked to the specific model, since this will depend on what the GNN has learned. For example, to classify a neighbor a GNN might be using some neighbors and the influence of these neighbors in the predictions would be its importances. Hence, for another GNN the important neighbors might be others since it might have learned different patterns.

The objective of this paper is following the third approach to explain how important is each neighbor for the node classification problem and measure it. By doing so, our goal is to understand how the GNN is using the neighbors to perform the classification of each node, increasing the trust of users in these type of models and helping them comprehend how is the model using that information. It is important to clarify here that the edges will not be explained, unlike in \cite{baldassarre2019explainability}, since in many cases the data does not have edge features, leaving the door open to continue this research in the future. Thus, following the trend of evaluating explainability methods mentioned above, there is currently no clear way of measuring which of the explainability techniques is better for a given task. Although there are several metrics published that try to measure different properties, as in \cite{ying2019gnnexplainer},  \cite{pope2019explainability}, \cite{sanchez-lengelingEvaluatingAttributionGraph2020}, whether or not the method accurately measures the relative importance of each neighbor in a node classification task, explained in Section~\ref{sec: metrics}.

The contributions of this paper can be summarized as below:
\begin{itemize}
    \item Creation of four metrics that allow for the first time to measure how well are explainability techniques inferring the importance of each neighbor for GNNs when classifying a node.
    \item Discovery that there are almost no differences in the explanations provided for GNNs when using gradient-based techniques (saliency maps and its variants), in contrast to computer vision domain, where the differences between the gradient-based methods are notorious.
    \item Discovery that for GNNs without self-loops many explainability techniques are not able to identify all the important neighbors, and therefore should be avoided by the users if the model they are constructing does not have self-loops.
    \item Adaptation of graph explainability techniques from computer vision and reformulation of PGExplainer, explained in Section~\ref{sec: pgexplainer}, with the objective of determining the importance of each neighbor for node classification.
\end{itemize}

\begin{figure}[h]
  \centering
  \begin{subfigure}{0.49\linewidth}
      \centering
      \includegraphics[width=\linewidth, height=0.8\linewidth]{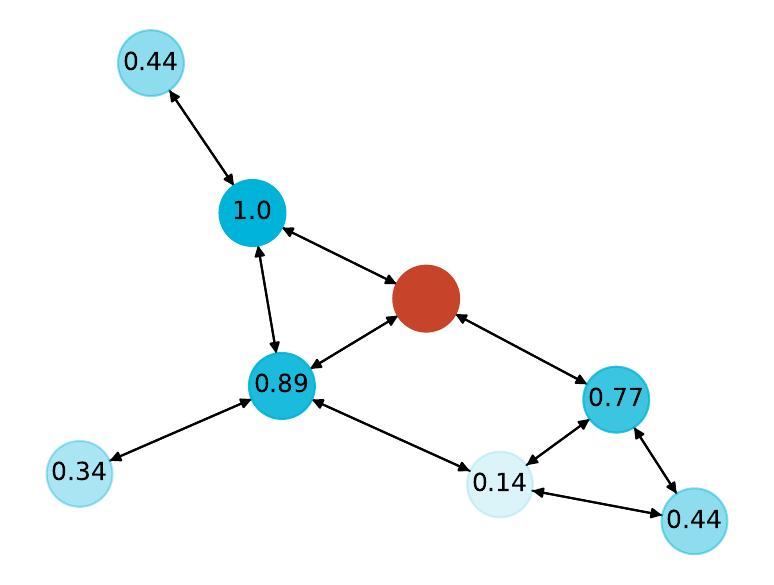}
      \caption{Saliency Map}
      \label{fig: saliency map}
  \end{subfigure}
  \begin{subfigure}{0.49\linewidth}
    \centering
    \includegraphics[width=\linewidth, height=0.8\linewidth]{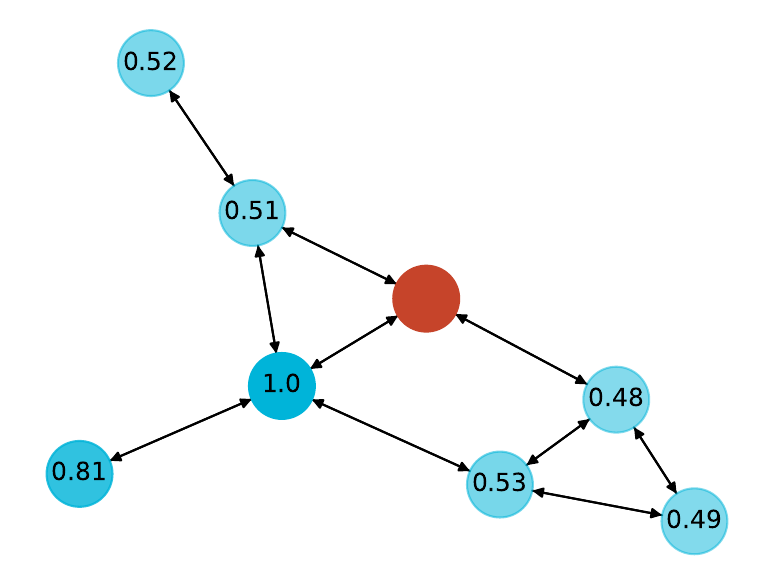}
    \caption{GNNExplainer}
    \label{fig: gnnexplainer}
  \end{subfigure}
  \caption{Explainability techniques to determine the importance of the neighbors (Saliency Map~\cite{simonyanDeepConvolutionalNetworks2014a} in Figure~\ref{fig: saliency map} and GNNExplainer~\cite{ying2019gnnexplainer} in Figure~\ref{fig: gnnexplainer}). Both are computed for node classification in Cora dataset with a GCN with self-loops. The red node is the node being classified and the others are the neighbors used for the classification. The importances are in the range [0-1] with 1 being the maximum.}
  \label{fig: explainability techniques}
\end{figure}

\section{Explainability methods}
\label{sec: explainability methods}
In this section the explainability methods to use will be detailed. Every method presented here is reformulated if necessary to give an importance score to each neighbor. This score will be a number that goes from 0, being the least important node for the classification, to 1, being the most important node. This concept is shown for two of the methods in Figure~\ref{fig: explainability techniques}. Due to limited space we have selected which explainability methods will be used based on popularity in the GNN domain, beside some techniques that have showed a great performance in computer vision and are adapted here to compare them with the rest of the methods. It is also important to note that here explanations, unlike in the case of the sub-graph, should be created with respect to the classification of a single node, since otherwise there would be no way of knowing the specific contributions to the classification of each node.

\subsection{Saliency Map}
\label{sec: saliency map}
This technique was introduced for the computer vision domain in \cite{simonyanDeepConvolutionalNetworks2014a}. It is based on the gradient computation, taking the absolute value of the gradient of the score of the predicted class $c$ with respect to the original image. Then, the saliency map adapted to GNNs would be the following:

\begin{equation} \label{eq: saliency map}
	M_{k}(\mathbf{X}_{i}) = \frac{1}{N} \sum_{j} \bigg | \frac{\partial \mathbf{y}_{kc}}{\partial \mathbf{X}_{ij}} \bigg |,
\end{equation}

where $\mathbf{X}$ is the node matrix, where $\mathbf{X}_{ij}$ is the $j$-th feature of neighbor $i$ (node $i$ is a neighbor of node $k$), similarly $\mathbf{X}_i$ denotes the feature vector of neighbor $i$, $N$ the total number of features, $\mathbf{y}_{kc}$ the classification score of the predicted class $c$ of node $k$ and $M_k(\mathbf{X})$ the saliency map of node $k$, respectively.

There have been attempts to adapt the technique before. In \cite{pope2019explainability} this technique was presented under the name of \textit{Contrastive gradient-based saliency maps}, but it only took the positive value of the gradients, which is in contrast to the specification in \cite{simonyanDeepConvolutionalNetworks2014a} that the absolute value is where the importance relies, regardless of the sign. In \cite{baldassarre2019explainability} it was also redefined, the only difference there is the reduction function used to get a final importance score for a specific node from the gradient for every feature. In the original paper, there was a gradient for every channel in the image. Since the authors wanted a 2D map, the maximum function was used to reduce the number of channels to one. In \cite{baldassarre2019explainability} the squared norm was used to have a unique value for the importance from the gradient for every feature in a specific node. Since in GNNs there could be many features, and in images there are only three channels, it is reasonable to use another reduction function. In this paper, the average is used because the norm can cause features with a high absolute value in the gradient to be over-represented. For the other gradient-based techniques used throughout the paper, the same reduction function will be used. However, for choosing an optimal reduction function an exhaustive analysis could be needed, which is out of the scope of this paper.

\subsection{Deconvnet and Guided-Backpropagation}
The Deconvnet technique was presented for the computer vision domain in \cite{zeilerVisualizingUnderstandingConvolutional2014a} that has not been adapted yet to GNNs domain. The only difference between the original implementation and the one presented here is the reduction function as in Section~\ref{sec: saliency map}.

The Guided-Backpropagation was presented for the computer vision domain in \cite{springenbergStrivingSimplicityAll2015b} and adapted in \cite{baldassarre2019explainability}. It is a combination of the saliency map and the deconvnet, where in a ReLU only the positive values of the gradients, after doing the original backpropagation in a ReLU, will be backpropagated.

\subsection{SmoothGrad}
This technique was presented for computer vision domain in \cite{smilkovSmoothGradRemovingNoise2017}, with the objective of reducing the noise of a saliency map. In \cite{sanchez-lengelingEvaluatingAttributionGraph2020} it was formulated a version of this technique using a combination of the gradients and the inputs, but here an adaptation of the original technique is presented. First, by following this definition for the saliency map before applying the reduction function:

%\pv{
\begin{equation}
	M_k(\mathbf{X}_{ij}) = \frac{\partial \mathbf{y}_{kc}}{\partial \mathbf{X}_{ij}},
\end{equation}

the equation of the SmoothGrad can be formulated in the following way:

\begin{equation}
	\hat{M}_{k}(\mathbf{X}_{i}) = \frac{1}{N} \sum_{j=1}^{N} \bigg | \frac{1}{n} \sum_{1}^{n} M_k(\mathbf{X}_{ij} + \mathcal{N}(0, \sigma^{2})) \bigg |,
\end{equation}

where $n$ represents the number of noise samples and $\mathcal{N}(0,\,\sigma^{2}))$ is Gaussian Noise with standard deviation $\sigma$.

\subsection{GNNExplainer}
This technique was presented in \cite{ying2019gnnexplainer}, for generating a mask for the edges and a mask for the features. These masks were learned to select an important sub-graph for the prediction, compounded from a subset of important nodes with a subset of important features. Since here the objective is to give an importance to each neighbor based on a node classification prediction, another approach has been followed. The implementation used here can be found in PyG library~\cite{feyFastGraphRepresentation2019}, where a mask is generated for the nodes instead of for the edges. Here the values of the mask before going through the sigmoid will be used as the importances. The reason for this is to avoid saturation of high values to better compare the importance of neighbors.

\subsection{PGExplainer}
\label{sec: pgexplainer}
This technique was proposed in \cite{luo2020parameterized} as in improvement over GNNExplainer. The technique learns a model $g_\psi$ that creates weights $w_{ij}$ for each edge $e_{ij}$ generating a mask that is defined by

\begin{equation}
	w_{ij} = MLP_\psi([z_i;z_j;z_v]),
\end{equation}

where $MLP_\psi$ is a multi-layer neural network parameterized with $\psi$ and $[\cdot;\cdot]$ is the concatenation operation. However this value must be mapped into a node importance to fulfill the objective of the problem addressed here. For that purpose, in this paper it has been chosen to assign the value from the edge $e_{ij}$ to the source node $i$. The reason for this choice is that if the edge is important in the direction $ij$ it is because the information coming from $i$ must be important according to the explainability technique, which seems like a reasonable assumption. 

% \subsection{Attention Explainer}

% The attention explainer based on GAT (Graph Attention Network) \cite{velickovic2018graph}. It is a graph neural architecture for graph-structure data. The model contains an attention mechanism. The mechanism adds a weighting factor to each link in order to take into account the significance of each neighbor. The method finds the edge with the highest attention score. This way it gets the most important neighbor. The definition of attention score is

% \[\alpha_{ij} = \frac{exp(LeakyReLU(\vec{a}^{T}{\boldsymbol{[\vec{W}}h_{i} || \vec{W}}h_{j}]))}{\sum_{k \in N_i}{exp(LeakyReLU(\vec{a}^{T}{\boldsymbol{[\vec{W}}h_{i} || \vec{W}}h_{k}]))}}\]

% where \( \vec{a} \in \mathbb{R}^{2F} \) is a weight vector (\(F\) is number of features.) and \(j \in N_i\) where \(j\) is some neighbors of node \(i\) and \(.^{T}\) represents transposition and \(||\) is the concatenation operation.

\section{Metrics}
\label{sec: metrics}
There may be several desirable properties that explainability techniques should have, as in \cite{pope2019explainability} or \cite{sanchez-lengelingEvaluatingAttributionGraph2020}, but
to be able to measure which technique is better for the problem addressed here it is needed to validate if the different explainability methods are identifying correctly how important is each neighbor for a GNN performing node classification. However, none of the metrics published in the literature can fulfill this objective. 

A common approach is to use metrics\footnote{Here we follow the same convention that has been used for a metric in the explainability literature as in \cite{pope2019explainability}, \cite{yuan2022explainability}, \cite{li2022explainability}, \cite{sanchez-lengelingEvaluatingAttributionGraph2020}, \cite{hookerBenchmarkInterpretabilityMethods2019a} or \cite{petsiukRISERandomizedInputa}, where the metric is not really a metric in the mathematical sense, but just a technique that tries to provide a score to each explainability method according to its performance in a specific task} such as the accuracy~\cite{ying2019gnnexplainer} or the Area Under the ROC Curve (AUC-ROC)~\cite{pope2019explainability}, where practitioners compare the results with ground-truth explanations from synthetic datasets. However, the GNN may learn to classify using a different set of importances for the neighbors than the one specified in the ground-truth explanations. Moreover, explainability methods were created because we cannot know what a neural networks has learned, as explained in \cite{simonyanDeepConvolutionalNetworks2014a}, \cite{smilkovSmoothGradRemovingNoise2017}, \cite{zhouLearningDeepFeatures2016} or \cite{selvarajuGradCAMVisualExplanations2017}, but using ground-truth explanations is precisely assuming that this information is already known.

Moreover, other popular metrics are fidelity~\cite{pope2019explainability} (new versions of this were introduced in \cite{yuan2022explainability}) and Explanation Confidence~\cite{li2022explainability}. Although they do not assume any previous knowledge, in both cases the metrics are designed to measure the ability of an explainability method to create a sub-graph (sub-set of nodes) that includes the nodes with an importance higher than a certain threshold. Therefore, these metrics quantify the performance of the explainability methods for minimizing the size of the whole graph while minimizing the loss in performance rather than for determining if they are identifying correctly how important is each neighbor for a GNN performing node classification. 

Furthermore, other popular metrics are Contrastivity~\cite{pope2019explainability} or Consistency~\cite{sanchez-lengelingEvaluatingAttributionGraph2020}, but they make certain assumptions that may not be true. First, Contrastivity assumes that explanations should differ between different classes. However, as pointed out in \cite{yuan2022explainability}, this might not be the case since different classes can have common patterns. And in the case of Consistency it assumes that high-performing models should have consistent explanations, which is not necessarily true as mentioned in \cite{yuan2022explainability}.

Therefore, four new metrics are proposed with the objective of measuring which technique is better to determine the importance of each neighbor for the GNN when classifying a node. The approach proposed here is to divide the objective into four sub-objectives, measured by the four metrics in a complementary way: 
\begin{itemize}
    \item Loyalty: Find which technique is better to identify the most important neighbors when they are highly important, i.e., removing them all will involve a change in the classification.
    \item Inverse loyalty: Find which technique is better to identify the least important neighbors.
    \item Loyalty probabilities: Find the most important neighbors on average.
    \item Inverse loyalty probabilities: Find the least important neighbors on average.
\end{itemize}

The metrics proposed here are inspired by the deletion metrics from the computer vision domain (\cite{petsiukRISERandomizedInputa} or \cite{hookerBenchmarkInterpretabilityMethods2019a}), since they are based on the concept of ordering the neighbors by importance and incrementally deleting them in fixed percentages (instead of deleting one by one it is done in fixed percentages to reduce the computation time). However, there are several differences from previous work and what is proposed here worth pointing out:

\begin{itemize}
    \item Here the comparison will be done between the original classification and the new classification, unlike in the previous proposed metrics where the accuracy was compared. In the case of the accuracy only correct examples are used, but testing should involve also the wrong examples since methods should also explain decisions when a model classifies incorrectly.
    \item Until now deletion metrics only use the descending order, starting by deleting the most important features, in this case neighbors. However, since these are accumulated metrics, the neighbors deleted at the end, the least important ones, may be affected to a large degree by having many neighbors deleted before them. This could be avoided by inverting the order, that is the concept showed by inverse loyalty and inverse loyalty probabilities. If these metrics are in fact giving new information, then the results should point out that some methods are the best ones to identify the most important neighbors (loyalty) but not the least (inverse loyalty), and that is precisely what can be observed in Section~\ref{sec: experiments and results}.
    \item In the case of deleting neighbors there is a different situation, since neighbor deletion may not change the classification since the importance may reside in the node itself. Therefore, it may be worth considering techniques are the best ones to identify the most and least important neighbors when the neighbors are highly important (loyalty and inverse loyalty). Alternatively, one may consider the most and least important neighbors on average, considering also non-highly important neighbors (loyalty and inverse loyalty probabilities).
\end{itemize}

All the metrics will measure the impact on the classification or its confidence of each node individually, as mentioned in Section~\ref{sec: explainability methods}, since otherwise scores could be due to several nodes being classified. Moreover, for all these metrics, as noted in Section~\ref{sec: explainability methods}, the node for the classification itself will be ignored since the objective is to measure the influence of the neighbors. Finally, it is important to note that a comparison with other metrics is not included since previous metrics were not measuring the influence of neighbors on node classification, making such a comparison impossible.

\subsection{Loyalty}
\label{sec: loyalty}
\begin{figure}[h]
  \centering
  \begin{subfigure}{0.49\linewidth}
      \centering
      \includegraphics[width=\linewidth, height=0.8\linewidth]{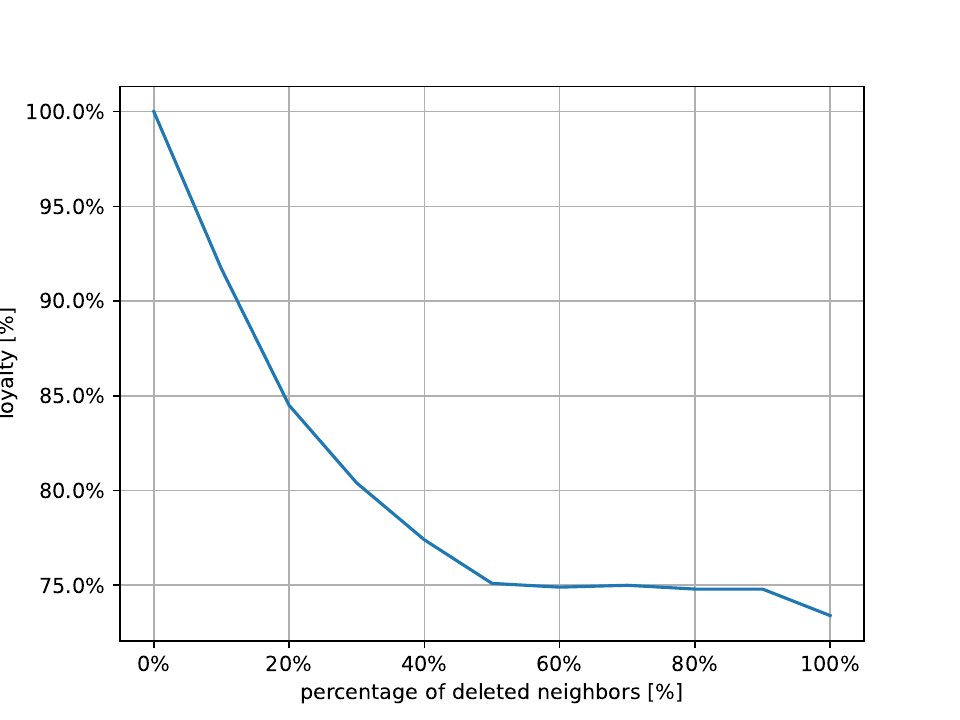}
      \caption{Loyalty}
      \label{fig: loyalty}
  \end{subfigure}
  \begin{subfigure}{0.49\linewidth}
    \centering
    \includegraphics[width=\linewidth, height=0.8\linewidth]{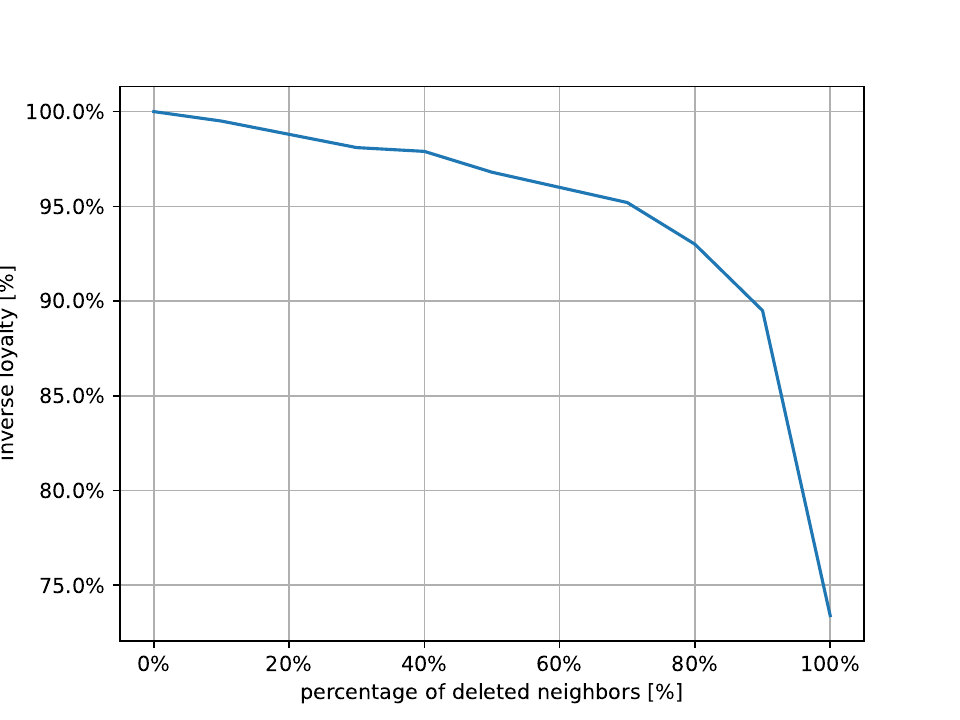}
    \caption{Inverse Loyalty}
    \label{fig: inverse loyalty}
  \end{subfigure}
  \caption{Loyalty and Inverse Loyalty of Saliency Map of GNNs with self-loops on Cora dataset}
  \label{fig: loyalty and inverse}
\end{figure}

For the loyalty the neighbors will be ordered based on their importance according to the explainability technique. For each node being classified, its neighbors with non-zero importance values will be sorted in decreasing order, i.e. starting with the most important neighbors according to the explainability method. Then, these neighbors will be progressively deleted by 10\% intervals. That way, first the 10\% most important will be deleted, then the 20\%, etc. For each node, classification for every percentage will be computed and compared with the original classification. It is important to note that despite what has been done previously for other metrics in the literature~\cite{yuan2022explainability}, here the comparison is between the classification of all nodes, regardless of whether the original classification was correct or incorrect. The reason for this is that even if the model made a mistake classifying a node, the model has used some neighbors for classifying it incorrectly, and by deleting the important neighbors this classification will be altered, even though the classification was originally wrong. Hence, for a specific percentage $k$ of neighbors deleted, we define the loyalty as:

\begin{equation}
	l_k = \frac{1}{N} \sum_{i=1}^{N} \mathds{1} (\hat{y}_{oi} = \hat{y}_{ki}),
\end{equation}

with $N$ being the number of total nodes being classified, $\hat{y}_{oi}$ the original classification of node $i$ and $\hat{y}_{ki}$ the classification of node $i$ with $k$ percentage of deleted neighbors. The progression of the loyalty throughout the different percentages of deleted neighbors can be observed in Figure~\ref{fig: loyalty}. The better the technique, the greater the drop in classification accuracy at the beginning, since when important nodes are deleted, the classification should be highly affected presented by a sharp decrease that becomes more gradual as less important neighbors are eliminated. In fact, this trend is contingent upon the graph's characteristics and the insights derived by the model. In scenarios where all neighbors share equal importance, the decline in error rate is expected to be uniform. In that special case, all techniques would yield identical results, regardless of the order of selection of important neighbors.

The 10\% intervals was chosen since it seems like a reasonable value to reduce the time complexity, that would be $O(NK)$, being $N$ the total number of nodes being classified and $K$ the total number of percentages to compute the metric.

\subsection{Inverse Loyalty}
This technique is computed the same way the loyalty is, with the exception being that the neighbors are sorted in ascending order for inverse loyalty. Therefore, the less important neighbors are identified correctly. This information was not visible in the loyalty because unimportant neighbors are deleted at the end, which can be affected by all the neighbors deleted before. Hence, by inverting the sorting order this information is clearly displayed. The progression of inverse loyalty throughout the different percentages of deleted neighbors can be visualized in Figure~\ref{fig: inverse loyalty}. In contrast to the loyalty metric, the best technique is the one with the smoother decrease at the beginning, where the least important neighbors are removed, that gets more steeper as we remove more important neighbors. Regarding the time complexity, it would be the same as the one showed for the loyalty.

\subsection{Loyalty Probabilities}
In Figure~\ref{fig: loyalty and inverse} one may notice that the metrics only decrease to a percentage around 75\%. That is because with a graph neural network that has self-loops it may be the case that even if all neighbors are deleted, the model still classifies correctly (explained in detail in Appendix~\ref{sec: loyaty with all neighbors deleted for gnns with self-loops}). However, it could be the case that the classification of a node does not change, but the confidence of that classification is lower. That can be measured if instead of computing the final classification, the class probabilities provided by the model are measured. For that purpose, we introduce the loyalty probabilities, constructed similarly to the loyalty but measuring the absolute difference of the probabilities (outputs of the model after a softmax layer) given to the original predicted class:

\begin{equation}
	l_k = \frac{1}{N} \sum_{i=1}^{N}  |(P(\hat{y}_i = \hat{y}_{oi} \mid \mathbf{G}=\mathbf{G_{ki}}) - (P(\hat{y}_i = \hat{y}_{oi} \mid \mathbf{G}=\mathbf{G_o})|,
\end{equation}

with $N$ being the number of total nodes being classified, $\hat{y}_{oi}$ the original classification of node $i$, $\hat{y}_i$ the classification of node $i$ with $\mathbf{G}$ as input, $\mathbf{G}_o$ the original input graph and 
$\mathbf{G}_{ki}$ the graph after deleting the $k$-th percentage of important nodes that influence the classification of node $i$ the most.

\begin{figure}[h]
  \centering
  \begin{subfigure}{0.49\linewidth}
      \centering
      \includegraphics[width=\linewidth, height=0.8\linewidth]{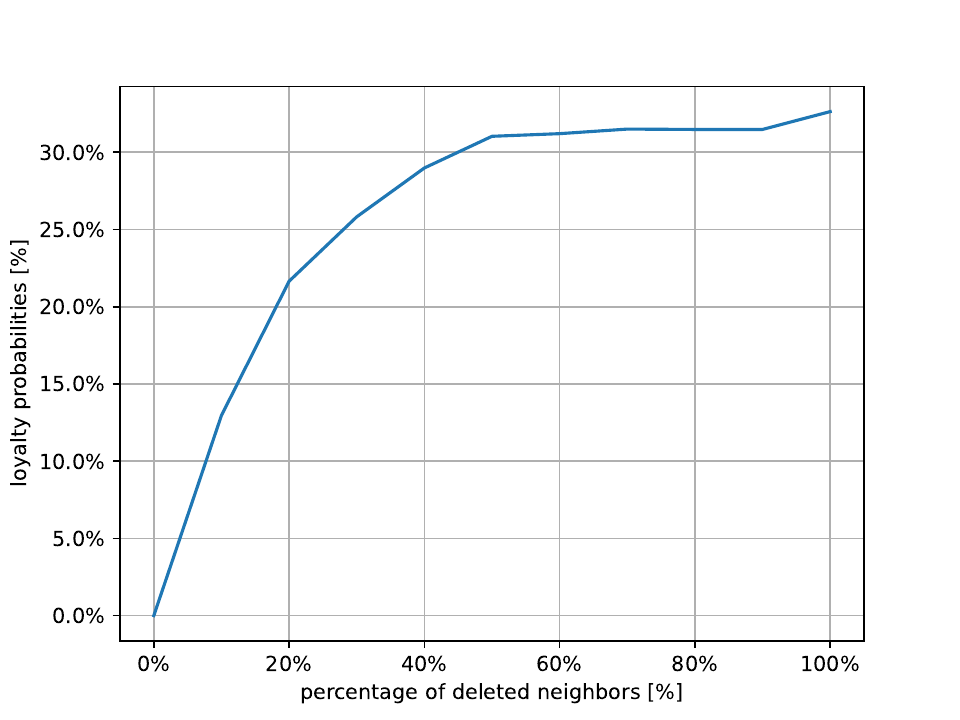}
      \caption{Loyalty Probabilities}
      \label{fig: loyalty probabilities}
  \end{subfigure}
  \begin{subfigure}{0.49\linewidth}
    \centering
    \includegraphics[width=\linewidth, height=0.8\linewidth]{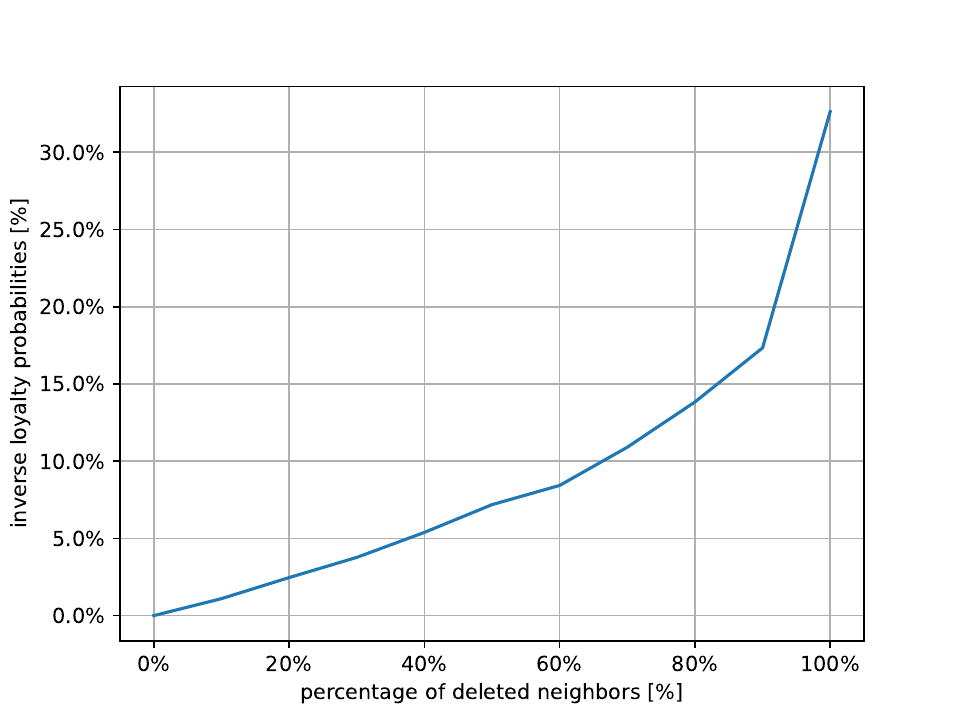}
    \caption{Inverse Loyalty Probabilities}
    \label{fig: inverse loyalty probabilities}
  \end{subfigure}
  \caption{Loyalty and Inverse Loyalty Probabilities of Saliency Map of GNNs with self-loops on Cora dataset}
  \label{fig: loyalty and inverse probablities}
\end{figure}

Hence, the evolution of the loyalty probabilities throughout the percentages of deleted neighbors can be found in Figure~\ref{fig: loyalty probabilities}. The better the technique, the sharper the increase should be at the beginning, since when important nodes are deleted the probability of the original predicted class should be highly affected, and the smoother at the end. Regarding the time complexity, it is the same as the one stated for the loyalty and inverse loyalty.

\subsection{Inverse Loyalty Probabilities}
Likewise, this metric is the version of the inverse loyalty with probabilities. The computation is exactly the same as the loyalty probabilities with the neighbors sorted in ascending order. The objective, similarly to the inverse loyalty, is to measure how well is the explainability method identifying unimportant neighbors. The evolution of the inverse loyalty probabilities throughout the percentages of deleted neighbors can be found in Figure~\ref{fig: inverse loyalty probabilities}. The better the technique, the smoother the increase should be at the beginning, since those neighbors are not important and in consequence the probabilities of the original predicted class should not change significantly, and the sharper at the end. Regarding the time complexity, it is the same as the one stated for the previous metrics.

\section{Experiments and Results}
\label{sec: experiments and results}
To test the different methods with the metrics explained in the previous section, two types of models and three different datasets will be used. The models used are a two layer GNN with a dropout~\cite{srivastavaDropoutSimpleWay2014} of 0.5 and a ReLU non-linearity between the GNN layers. The first model will have two GCN~\cite{kipfSemiSupervisedClassificationGraph2017} layers and the second two GAT layers (the second version pesented in \cite{brodyHowAttentiveAre2021}). Then, the datasets used are the citation network datasets~\cite{yangRevisitingSemisupervisedLearning2016}, Cora, CiteSeer and PubMed, since they are commonly used for node classification. That way it will be shown that the approach presented here can be used with any dataset, since no ground truth explanations are necessary. The explainability techniques are evaluated only in the test data. Its performance can be checked in Appendix~\ref{sec: performance on test set}.

Even though the visualization of the evolution of the metrics throughout the different percentages of deleted neighbors can provide useful information, a way of quantifying it in a single number to compare different techniques is needed. For that purpose, the Area Under the Curve (AUC) of the visualizations showed in the last Section will be computed. Hence, the best technique is the one with minimum AUC for loyalty and maximum AUC for inverse loyalty, since in the former a sharper decrease is better at the beginning and a smoother evolution at the end, and in the latter the opposite, as it was explained in Section~\ref{sec: metrics}. These metrics can be visualized in Table~\ref{tab: auc loyalty and inverse loyalty}.

\begin{table}[h]
	\centering
	\caption{AUC Loyalty (L) and Inverse (I) Loyalty}
	\label{tab: auc loyalty and inverse loyalty}
	\scalebox{0.9}{
    	\begin{tabular}{cccccccccccccc}
    		\toprule
    		 \multirow{3.5}{*}{Self-Loops} & \multirow{3.5}{*}{Method} & \multicolumn{4}{c}{Cora} & \multicolumn{4}{c}{CiteSeer} & \multicolumn{4}{c}{PubMed}  \\
    		\cmidrule(lr){3-6} \cmidrule(lr){7-10} \cmidrule(lr){11-14}
    	    & & \multicolumn{2}{c}{GCN} & \multicolumn{2}{c}{GAT} & \multicolumn{2}{c}{GCN} & \multicolumn{2}{c}{GAT} & \multicolumn{2}{c}{GCN} & \multicolumn{2}{c}{GAT} \\
    % 		\cmidrule(lr){3-4} \cmidrule(lr){5-6}  
    		& & L & I & L & I & L & I & L & I & L & I & L & I \\
    		\midrule
    		\multirow{6}{*}{With} 
    		& Saliency Map & 0.80  & 0.95 & 0.67 & \bf{0.94} & 0.86 & 0.86 & 0.81 & 0.94 & 0.83 & 0.96 & 0.84 & \bf{0.97} \\
    		& Smoothgrad & 0.80  & 0.95 & 0.70  & 0.89 & 0.87 & 0.96 & 0.81 & 0.93 & 0.83 & 0.96 & 0.84 & 0.96 \\
    		& Deconvnet & 0.79  & 0.95 & 0.67  & \bf{0.94} & 0.86 & 0.86 & 0.81 & 0.94 & 0.83 & 0.96 & 0.84 & \bf{0.97} \\
    		& Guided Backprop & 0.79  & 0.95 & 0.67  & \bf{0.94} & 0.86 & 0.86 & 0.81 & 0.94 & 0.83 & 0.96 & 0.84 & \bf{0.97} \\
    		& GNNExplainer & \bf{0.74}  & \bf{0.97}  & \bf{0.64}  & \bf{0.94} & \bf{0.83} & \bf{0.97} & \bf{0.78} & \bf{0.96} & \bf{0.79} & \bf{0.97} & \bf{0.81} & \bf{0.97} \\
    		& PGExplainer & 0.88 & 0.88  & 0.75  & 0.86 & 0.91 & 0.90 & 0.85 & 0.89 & 0.90 & 0.86 & 0.88 & 0.91 \\
    		\midrule
    	    \multirow{6}{*}{Without}
    		& Saliency Map & 0.81 & 0.90 & 0.58 & 0.77 & 0.89  & 0.93 & 0.78 & 0.70 & 0.89 & 0.9 & 0.58 & 0.81 \\
    		& Smoothgrad & 0.81 & 0.90  & 0.57 & 0.78 & 0.89 & 0.92 & 0.77 & 0.71 & 0.89 & 0.93 & \bf{0.54} & \bf{0.85} \\
    		& Deconvnet & 0.81 & 0.90 & 0.58 & 0.79 & 0.90 & 0.92 & 0.79 & 0.69 & 0.89 & 0.93 & 0.58 & 0.80 \\
    		& Guided Backprop & 0.81 & 0.90 & 0.58 & 0.79 & 0.90 & 0.92 & 0.79 & 0.69 & 0.89 & 0.93 & 0.58 & 0.80 \\
    		& GNNExplainer & \bf{0.74} & \bf{0.94} & \bf{0.56} & \bf{0.80} & 0.86 & \bf{0.96} & \bf{0.74} & \bf{0.76} & 0.77 & \bf{0.97} & 0.7 & 0.71 \\
    		& PGExplainer & 0.76 & 0.73 & 0.66 & 0.73 & \bf{0.80} & 0.73 & 0.75 & 0.75 & \bf{0.74} & 0.72 & 0.67 & 0.74 \\
    		\bottomrule
    	\end{tabular}
    }
\end{table}

First, by analyzing the results of the models with self-loops, it can be observed that the best explainability technique (highlighted in bold) in all cases is the GNNExplainer. Moreover, all gradient-based methods (saliency map, smoothgrad, deconvnet and guided backprop) surprisingly have almost the same result, in contrast to computer vision domain, where the differences between the gradient-based methods are notorious. A plausible explanation for this is that the differences in gradient-based methods accumulate throughout the layers of the models. Since computer vision models tend to be really deep, the differences are greater, while GNNs tend to be shallow and produce much smaller variations. Even though this is an exciting topic, it is out of the scope of the paper, so we leave it for future research. Lastly, PGExplainer is the worst in almost all cases. However, without the self-loops, the mentioned behaviour changes, with the PGExplainer becoming the second best technique after the GNNExplainer. 

To check if this behavior holds with the probability-based metrics Table~\ref{tab: auc loyalty and inverse loyalty probabilities} can be visualized. Since the loyalty probabilities indicate that an explainability method is better when the increase is sharper at the beginning and smoother at the end, and the inverse loyalty probabilities is the opposite, for the former the best technique is the one that maximizes the AUC and for the latter the one that minimizes it.

\begin{table}[h]
	\centering
	\caption{AUC Loyalty (L) and Inverse (I) Loyalty Probabilities}
	\label{tab: auc loyalty and inverse loyalty probabilities}
	\scalebox{0.9}{
    	\begin{tabular}{cccccccccccccc}
    		\toprule
    		 \multirow{3.5}{*}{Self-Loops} & \multirow{3.5}{*}{Method} & \multicolumn{4}{c}{Cora} & \multicolumn{4}{c}{CiteSeer} & \multicolumn{4}{c}{PubMed}  \\
    		\cmidrule(lr){3-6} \cmidrule(lr){7-10} \cmidrule(lr){11-14}
    	    & & \multicolumn{2}{c}{GCN} & \multicolumn{2}{c}{GAT} & \multicolumn{2}{c}{GCN} & \multicolumn{2}{c}{GAT} & \multicolumn{2}{c}{GCN} & \multicolumn{2}{c}{GAT} \\
    % 		\cmidrule(lr){3-4} \cmidrule(lr){5-6}  
    		& & L & I & L & I & L & I & L & I & L & I & L & I \\
    		\midrule
    		\multirow{6}{*}{With} 
    		& Saliency Map & 0.26 & \bf{0.09} & 0.40 & 0.08 & 0.17 & \bf{0.07} & 0.24 & \bf{0.07} & \bf{0.22} & \bf{0.06} & 0.20 & \bf{0.04} \\
    		& SmoothGrad & 0.26 & \bf{0.09} & 0.37 & 0.14 & 0.17 & \bf{0.07} & 0.23 & 0.08 & \bf{0.22} & \bf{0.06} & 0.19 & 0.05\\
    		& Deconvnet & 0.26 & \bf{0.09} & 0.40 &  \bf{0.07} & 0.17 & \bf{0.07} & 0.24 & \bf{0.07} & \bf{0.22} & \bf{0.06} & 0.20 & \bf{0.04}\\
    		& Guided Backprop & 0.26 & \bf{0.09} & 0.40 & \bf{0.07} & 0.17 & \bf{0.07} & 0.24 & \bf{0.07} & \bf{0.22} & \bf{0.06} & 0.20 & \bf{0.04}\\
    		& GNNExplainer & \bf{0.28} & 0.15 & \bf{0.41} & 0.10 & \bf{0.18} & 0.11 & \bf{0.25} & 0.08 & \bf{0.22} & 0.10 & \bf{0.21} & 0.06\\
    		& PGExplainer & 0.18 & 0.17 & 0.31 & 0.19 & 0.12 & 0.13 & 0.19 & 0.13 & 0.13 & 0.17 & 0.15 & 0.11\\
    		\midrule
    	    \multirow{6}{*}{Without}
    		& Saliency Map & 0.24 & \bf{0.14} & 0.46 & 0.24 & 0.13 & \bf{0.09} & 0.24 & 0.28 & 0.15 & \bf{0.10} & 0.40 & 0.18\\
    		& SmoothGrad & 0.24 & \bf{0.14} & 0.45 & 0.26 & 0.13 & \bf{0.09} & 0.24 & 0.27 & 0.15 & \bf{0.10} & \bf{0.43} & \bf{0.14}\\
    		& Deconvnet & 0.24 & \bf{0.14} & 0.46 & \bf{0.23} & 0.13 & \bf{0.09} & 0.24 & 0.28 & 0.15 & \bf{0.10} & 0.39 & 0.19\\
    		& Guided Backprop & 0.24 & \bf{0.14} & 0.46 & \bf{0.23} & 0.13 & \bf{0.09} & 0.24 & 0.28 & 0.15 & \bf{0.10} & 0.39 & 0.19\\
    		& GNNExplainer & \bf{0.28} & 0.17 & \bf{0.47} & \bf{0.23} & \bf{0.14} & 0.11 & \bf{0.26} & \bf{0.24} & 0.23 & 0.13 & 0.32 & 0.26\\
    		& PGExplainer & 0.26 & 0.30 & 0.39 & 0.30 & 0.20 & 0.26 & \bf{0.26} & \bf{0.24} & \bf{0.24} & 0.28 & 0.34 & 0.23\\
    		\bottomrule
    	\end{tabular}
    }
\end{table}

Looking at the results for the probabilities metrics, the gradient-based methods perform even better than the GNNExplainer. Moreover, there is still almost no difference between these gradient-based techniques. Therefore, it might be the case that the GNNExplainer is better for capturing the importance of the neighbors when the impact of them is higher (classification of the node changes) but gradient-based methods are better in general when the classification may not change, but the confidence of the model does. Although this is out of the scope of this paper, a plausible explanation for this could be that by training a mask with gradient descent local minima are found when there are not high impact neighbors. In addition, PGExplainer is still the worst when there are self-loops and without them its results are similar to the ones obtained with the other methods. 

% \pv{Can we give here some hints or intuitions why we get these results? 

% We should also make it clear that these numbers reflect the performance of determining the important or unimportant nodes, not the performance of the explainers themselves.}

\begin{figure}[h]
  \centering
  \begin{subfigure}{0.49\linewidth}
      \centering
      \includegraphics[width=\linewidth, height=0.8\linewidth]{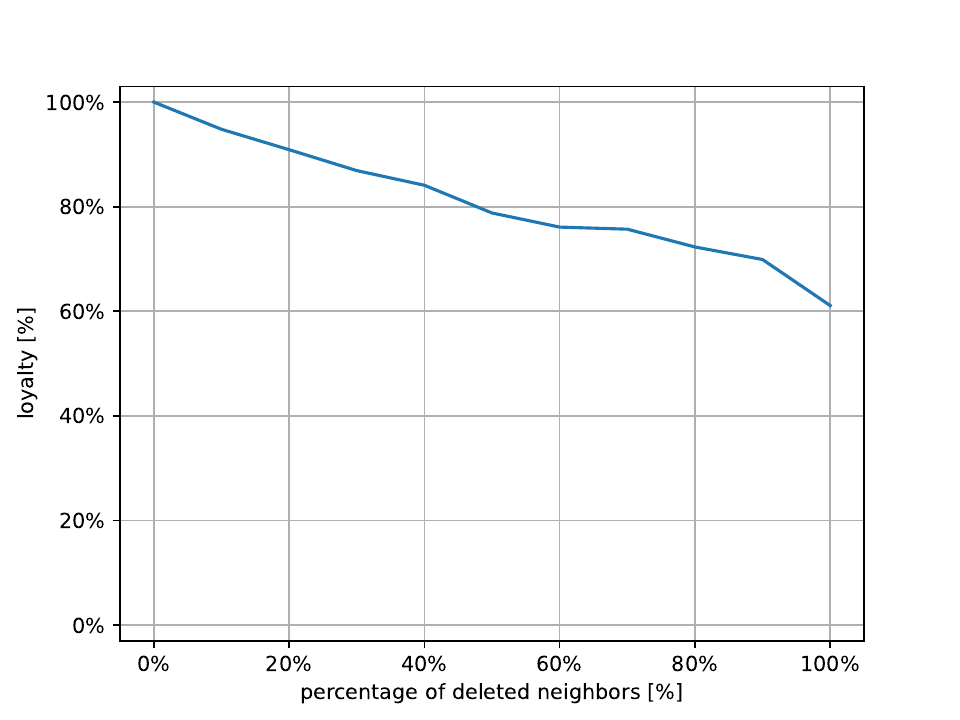}
      \caption{Saliency Map}
  \end{subfigure}
  \begin{subfigure}{0.49\linewidth}
    \centering
    \includegraphics[width=\linewidth, height=0.8\linewidth]{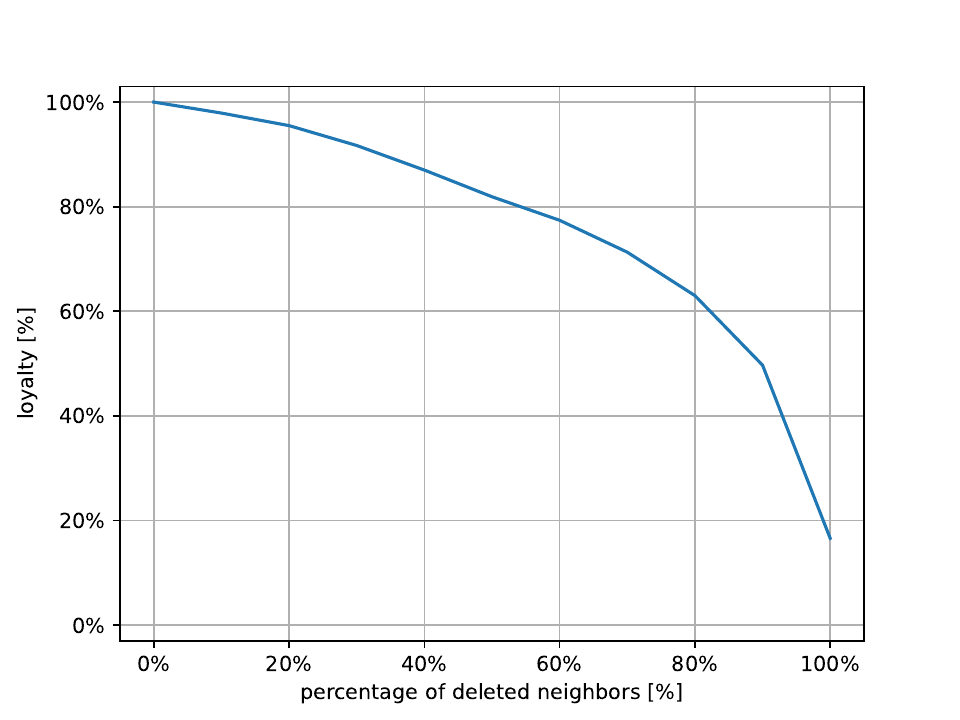}
    \caption{PGExplainer}
    \label{fig: loyalty without self-loops pg}
  \end{subfigure}
  \caption{Loyalty without self-loops}
  \label{fig: loyalty without self-loops}
\end{figure}

To explore in detail the difference between the explainability results with and without self-loops Figure~\ref{fig: loyalty without self-loops} has been constructed. Based on the visualization it can be observed that the saliency map is not able to correctly attribute importance to all the neighbors. In the visualizations showed in Section~\ref{sec: metrics} the loyalty and inverse loyalty only decrease to a certain point around 75\% because with self-loops, even if all neighbors are deleted, the metrics only decrease to a certain point (proof for that can be found in Appendix~\ref{sec: loyaty with all neighbors deleted for gnns with self-loops}). However, in Figure~\ref{fig: loyalty without self-loops} it can be observed how without self-loops, with the PGExplainer the loyalty decreases to a point around 20\%, while with the saliency map it is stuck at 60\%. Around 20\% in Cora dataset is in fact approximately the level the loyalty of a good technique should decrease, since without self-loops the classification is almost random (depends on bias parameter) once every neighbor has been deleted. To check this phenomenon in every dataset, Table~\ref{tab: loyalty with all neighbors deleted} was constructed, where the loyalties when all neighbors with non-zero importance (according to each explainability technique) are deleted and compared with the loyalties when all neighbors are deleted for GNNs without self-loops. Therefore, a technique that is able to detect all important neighbors should have the same performance as the last row in Table~\ref{tab: loyalty with all neighbors deleted}, and only PGExplainer accomplish that. The reason for this is that the other techniques use the gradient for their computation (GNNExplainer uses it also for training the mask) and without self-loops gradients are not backpropagated to the node itself only to the neighbors, which can cause situations such as the one explained in Appendix~\ref{sec: not finding neighbors} that prevent techniques from discovering all the important neighbors. It is also worth mentioning that the results of the PGExplainer are inadequate since it is offering similar results as the others in spite of being able to discover all the neighbors. Its poor performance can also be observed in the shape of Figure~\ref{fig: loyalty without self-loops pg}, which is the opposite as the one it should be, as explained in Section~\ref{sec: metrics}. This is consistent with what was found in \cite{holdijkReParameterizedExplainer2021} and \cite{yuan2022explainability}, where the performance of the technique was substantially worse than the one originally presented in \cite{luo2020parameterized}.

\begin{table}[h]
	\centering
	\caption{Loyalty with all Neighbors deleted for GNNs without self-loops}
	\label{tab: loyalty with all neighbors deleted}
	\scalebox{0.9}{
    	\begin{tabular}{ccccccccccccc}
    		\toprule
    		\multirow{2.5}{*}{Method} & \multicolumn{2}{c}{Cora} & \multicolumn{2}{c}{CiteSeer} & \multicolumn{2}{c}{PubMed}  \\
    		\cmidrule(lr){2-3} \cmidrule(lr){4-5} \cmidrule(lr){6-7}
    	    & GCN & GAT & GCN & GAT & GCN & GAT \\
    		\midrule
    		Saliency Map & 0.61 & 0.22 & 0.75 & 0.31 & 0.76 & 0.29 &\\
    		Smoothgrad & 0.61 & 0.22 & 0.75 & 0.31 & 0.76 & 0.29 &\\
    		Deconvnet & 0.61 & 0.24 & 0.75 & 0.31 & 0.76 & 0.29 &\\
    		Guided Backprop & 0.61 & 0.24 & 0.75 & 0.31 & 0.76 & 0.29 &\\
    		GNNExplainer & 0.61 & 0.22 & 0.75 & 0.32 & 0.76 & 0.29 &\\
    		PGExplainer & 0.17 & 0.19 & 0.20 & 0.20 & 0.43 & 0.29 &\\
    		Without Neighbors & 0.17 & 0.19 & 0.20 & 0.20 & 0.43 & 0.29 &\\
    		\bottomrule
    	\end{tabular}
    }
\end{table}

\section{Conclusion and Future Work}
This paper addresses the problem of determining the importance of each neighbor for a GNN when classifying a node, specifically how to evaluate explainability techniques for this specific task. This will allow researchers and engineers to understand how important are the neighbors for the model.

In order to evaluate the mentioned methods, four new metrics have been presented. First, the loyalty metric is established to tests whether a method identifies correctly the most important neighbors, while the inverse loyalty is defined to evaluate whether least important neighbors are properly detected. However, since these two metrics only focus on neighbors when they are highly important, i.e., removing them all will involve a change in the classification, two other new metrics are introduced. The probabilistic versions of loyalty and inverse loyalty measure changes in the probabilities rather than the final prediction, being able to capture the influence of all neighbors, while the previous versions presented focus only on the highly important ones. This brings up two types of evaluations, one focused on cases where neighbors are so important that they can change the classification, and the other for a more general analysis.

Based on the comparison between the metrics results of the different explainability methods, the first discovery has been that, although in computer vision there is a great difference between the outputs of gradient-based techniques (as it is shown in \cite{adebayoSanityChecksSaliency2018a}), for GNNs they seem to provide almost the same results. Moreover, these gradient-based methods are the best ones at measuring the neighbor importance in general, since they have better AUC for loyalty and inverse loyalty probabilities. However, for the highly important neighbors neighbors GNNExplainer provides the best results. Moreover, PGExplainer is the worst by far, being consistent with what was found in \cite{holdijkReParameterizedExplainer2021} and \cite{yuan2022explainability}.

In addition, we have shown that there is a significant difference in the performance of many explainability methods when self-loops are not present. Specifically, the methods that use the gradients, either because they are gradient-based techniques that use it for computing the importance or because they need it for training a mask as in the GNNExplainer, are not able to identify many important neighbors when there are no self-loops. Therefore, these discoveries show that if a researcher builds a GNN without self-loops, then gradient-based explainability techniques should not be used to understand the importance of the neighbors.

As future work we leave the door open to research about methods that could mitigate the deficiencies of techniques that use the gradients when there are no self-loops and to find why all gradient-based methods behave in a similar manner.

% \section*{Author Contributions}
% \textbf{Oscar Llorente:} Conceptualization, Methodology, Investigation, Software, Writing - Original Draft and Supervision

\section*{Acknowledgements}
We thank our colleagues Javier Albert and Víctor Buenestado for their insightful comments and critical feedback. We thank Ericsson Cognitive Network Solutions, Ericsson Global Artificial Intelligence Accelerator and Ericsson Research for supporting this research.

% For natbib users:

\bibliographystyle{splncs04.bst}
\bibliography{main.bib}

\begin{thebibliography}{10}
\providecommand{\url}[1]{\texttt{#1}}
\providecommand{\urlprefix}{URL }
\providecommand{\doi}[1]{https://doi.org/#1}

\bibitem{adebayoSanityChecksSaliency2018a}
Adebayo, J., Gilmer, J., Muelly, M., Goodfellow, I., Hardt, M., Kim, B.: Sanity
  {{Checks}} for {{Saliency Maps}}. In: Advances in {{Neural Information
  Processing Systems}}. vol.~31. {Curran Associates, Inc.} (2018)

\bibitem{amannExplainabilityArtificialIntelligence2020}
Amann, J., Blasimme, A., Vayena, E., Frey, D., Madai, V.I., {the Precise4Q
  consortium}: Explainability for artificial intelligence in healthcare: A
  multidisciplinary perspective. BMC Medical Informatics and Decision Making
  \textbf{20}(1), ~310 (Nov 2020). \doi{10.1186/s12911-020-01332-6}

\bibitem{baldassarre2019explainability}
Baldassarre, F., Azizpour, H.: Explainability techniques for graph
  convolutional networks. In: International Conference on Machine Learning
  (ICML) Workshops, 2019 Workshop on Learning and Reasoning with
  Graph-Structured Representations (2019),
  \url{https://graphreason.github.io/papers/25.pdf}

\bibitem{brodyHowAttentiveAre2021}
Brody, S., Alon, U., Yahav, E.: How {{Attentive}} are {{Graph Attention
  Networks}}? In: International {{Conference}} on {{Learning Representations}}
  (Oct 2021)

\bibitem{brownLanguageModelsAre2020c}
Brown, T., Mann, B., Ryder, N., Subbiah, M., Kaplan, J.D., Dhariwal, P.,
  Neelakantan, A., Shyam, P., Sastry, G., Askell, A., Agarwal, S.,
  {Herbert-Voss}, A., Krueger, G., Henighan, T., Child, R., Ramesh, A.,
  Ziegler, D., Wu, J., Winter, C., Hesse, C., Chen, M., Sigler, E., Litwin, M.,
  Gray, S., Chess, B., Clark, J., Berner, C., McCandlish, S., Radford, A.,
  Sutskever, I., Amodei, D.: Language {{Models}} are {{Few-Shot Learners}}. In:
  Advances in {{Neural Information Processing Systems}}. vol.~33, pp.
  1877--1901. {Curran Associates, Inc.} (2020)

\bibitem{dikmenTrustAutonomousVehicles2017}
Dikmen, M., Burns, C.: Trust in autonomous vehicles: {{The}} case of {{Tesla
  Autopilot}} and {{Summon}}. In: 2017 {{IEEE International Conference}} on
  {{Systems}}, {{Man}}, and {{Cybernetics}} ({{SMC}}). pp. 1093--1098 (Oct
  2017). \doi{10.1109/SMC.2017.8122757}

\bibitem{fanGraphNeuralNetworks2019}
Fan, W., Ma, Y., Li, Q., He, Y., Zhao, E., Tang, J., Yin, D.: Graph {{Neural
  Networks}} for {{Social Recommendation}}. In: The {{World Wide Web
  Conference}}. pp. 417--426. {{WWW}} '19, {Association for Computing
  Machinery}, {New York, NY, USA} (May 2019). \doi{10.1145/3308558.3313488}

\bibitem{feyFastGraphRepresentation2019}
Fey, M., Lenssen, J.E.: Fast {{Graph Representation Learning}} with {{PyTorch
  Geometric}}. CoRR  \textbf{abs/1903.02428} (2019)

\bibitem{holdijkReParameterizedExplainer2021}
Holdijk, L., Boon, M., Henckens, S., de~Jong, L.: [{{Re}}] {{Parameterized
  Explainer}} for {{Graph Neural Network}}. In: {{ML Reproducibility
  Challenge}} 2020 (Jan 2021)

\bibitem{hookerBenchmarkInterpretabilityMethods2019a}
Hooker, S., Erhan, D., Kindermans, P.J., Kim, B.: A {{Benchmark}} for
  {{Interpretability Methods}} in {{Deep Neural Networks}}. In: Advances in
  {{Neural Information Processing Systems}}. vol.~32. {Curran Associates, Inc.}
  (2019)

\bibitem{izadiOptimizationGraphNeural2020}
Izadi, M.R., Fang, Y., Stevenson, R., Lin, L.: Optimization of {{Graph Neural
  Networks}} with {{Natural Gradient Descent}}. In: 2020 {{IEEE International
  Conference}} on {{Big Data}} ({{Big Data}}). pp. 171--179 (Dec 2020).
  \doi{10.1109/BigData50022.2020.9378063}

\bibitem{jumperHighlyAccurateProtein2021}
Jumper, J., Evans, R., Pritzel, A., Green, T., Figurnov, M., Ronneberger, O.,
  Tunyasuvunakool, K., Bates, R., {\v Z}{\'i}dek, A., Potapenko, A., Bridgland,
  A., Meyer, C., Kohl, S.A.A., Ballard, A.J., Cowie, A., {Romera-Paredes}, B.,
  Nikolov, S., Jain, R., Adler, J., Back, T., Petersen, S., Reiman, D., Clancy,
  E., Zielinski, M., Steinegger, M., Pacholska, M., Berghammer, T., Bodenstein,
  S., Silver, D., Vinyals, O., Senior, A.W., Kavukcuoglu, K., Kohli, P.,
  Hassabis, D.: Highly accurate protein structure prediction with
  {{AlphaFold}}. Nature  \textbf{596}(7873),  583--589 (Aug 2021).
  \doi{10.1038/s41586-021-03819-2}

\bibitem{kipfSemiSupervisedClassificationGraph2017}
Kipf, T.N., Welling, M.: Semi-{{Supervised Classification}} with {{Graph
  Convolutional Networks}}. In: 5th {{International Conference}} on {{Learning
  Representations}}, {{ICLR}} 2017, {{Toulon}}, {{France}}, {{April}} 24-26,
  2017, {{Conference Track Proceedings}}. {OpenReview.net} (2017)

\bibitem{kirillovSegmentAnything2023}
Kirillov, A., Mintun, E., Ravi, N., Mao, H., Rolland, C., Gustafson, L., Xiao,
  T., Whitehead, S., Berg, A.C., Lo, W.Y., Doll{\'a}r, P., Girshick, R.:
  Segment {{Anything}} (Apr 2023). \doi{10.48550/arXiv.2304.02643}, comment:
  Project web-page: https://segment-anything.com

\bibitem{li2022explainability}
Li, P., Yang, Y., Pagnucco, M., Song, Y.: Explainability in graph neural
  networks: An experimental survey (2022)

\bibitem{luo2020parameterized}
Luo, D., Cheng, W., Xu, D., Yu, W., Zong, B., Chen, H., Zhang, X.:
  Parameterized explainer for graph neural network. Advances in neural
  information processing systems  \textbf{33},  19620--19631 (2020)

\bibitem{molnarInterpretableMachineLearning}
Molnar, C.: Interpretable Machine Learning. Independently published, 2 edn.
  (2022), \url{https://christophm.github.io/interpretable-ml-book}

\bibitem{petsiukRISERandomizedInputa}
Petsiuk, V., Das, A., Saenko, K.: Rise: Randomized input sampling for
  explanation of black-box models. ArXiv  \textbf{abs/1806.07421} (2018),
  \url{https://api.semanticscholar.org/CorpusID:49324724}

\bibitem{pope2019explainability}
Pope, P.E., Kolouri, S., Rostami, M., Martin, C.E., Hoffmann, H.:
  Explainability methods for graph convolutional neural networks. In:
  Proceedings of the IEEE/CVF conference on computer vision and pattern
  recognition. pp. 10772--10781 (2019)

\bibitem{reiserGraphNeuralNetworks2022}
Reiser, P., Neubert, M., Eberhard, A., Torresi, L., Zhou, C., Shao, C., Metni,
  H., {van Hoesel}, C., Schopmans, H., Sommer, T., Friederich, P.: Graph neural
  networks for materials science and chemistry. Communications Materials
  \textbf{3}(1),  1--18 (Nov 2022). \doi{10.1038/s43246-022-00315-6}

\bibitem{sanchez-lengelingEvaluatingAttributionGraph2020}
{Sanchez-Lengeling}, B., Wei, J., Lee, B., Reif, E., Wang, P., Qian, W.,
  McCloskey, K., Colwell, L., Wiltschko, A.: Evaluating {{Attribution}} for
  {{Graph Neural Networks}}. In: Advances in {{Neural Information Processing
  Systems}}. vol.~33, pp. 5898--5910. {Curran Associates, Inc.} (2020)

\bibitem{selvarajuGradCAMVisualExplanations2017}
Selvaraju, R.R., Cogswell, M., Das, A., Vedantam, R., Parikh, D., Batra, D.:
  Grad-{{CAM}}: {{Visual Explanations}} from {{Deep Networks}} via
  {{Gradient-Based Localization}}. In: 2017 {{IEEE International Conference}}
  on {{Computer Vision}} ({{ICCV}}). pp. 618--626 (Oct 2017).
  \doi{10.1109/ICCV.2017.74}

\bibitem{shenGraphNeuralNetworks2021}
Shen, Y., Shi, Y., Zhang, J., Letaief, K.B.: Graph {{Neural Networks}} for
  {{Scalable Radio Resource Management}}: {{Architecture Design}} and
  {{Theoretical Analysis}}. IEEE Journal on Selected Areas in Communications
  \textbf{39}(1),  101--115 (Jan 2021). \doi{10.1109/JSAC.2020.3036965}

\bibitem{shiMaskedLabelPrediction2021}
Shi, Y., Huang, Z., Feng, S., Zhong, H., Wang, W., Sun, Y.: Masked {{Label
  Prediction}}: {{Unified Message Passing Model}} for {{Semi-Supervised
  Classification}}. In: Twenty-{{Ninth International Joint Conference}} on
  {{Artificial Intelligence}}. vol.~2, pp. 1548--1554 (Aug 2021).
  \doi{10.24963/ijcai.2021/214}

\bibitem{simonyanDeepConvolutionalNetworks2014a}
Simonyan, K., Vedaldi, A., Zisserman, A.: Deep {{Inside Convolutional
  Networks}}: {{Visualising Image Classification Models}} and {{Saliency
  Maps}}. In: Bengio, Y., LeCun, Y. (eds.) 2nd {{International Conference}} on
  {{Learning Representations}}, {{ICLR}} 2014, {{Banff}}, {{AB}}, {{Canada}},
  {{April}} 14-16, 2014, {{Workshop Track Proceedings}} (2014)

\bibitem{smilkovSmoothGradRemovingNoise2017}
Smilkov, D., Thorat, N., Kim, B., Vi{\'e}gas, F.B., Wattenberg, M.:
  {{SmoothGrad}}: Removing noise by adding noise. CoRR  \textbf{abs/1706.03825}
  (2017)

\bibitem{springenbergStrivingSimplicityAll2015b}
Springenberg, J.T., Dosovitskiy, A., Brox, T., Riedmiller, M.A.: Striving for
  {{Simplicity}}: {{The All Convolutional Net}}. In: Bengio, Y., LeCun, Y.
  (eds.) 3rd {{International Conference}} on {{Learning Representations}},
  {{ICLR}} 2015, {{San Diego}}, {{CA}}, {{USA}}, {{May}} 7-9, 2015, {{Workshop
  Track Proceedings}} (2015)

\bibitem{srivastavaDropoutSimpleWay2014}
Srivastava, N., Hinton, G., Krizhevsky, A., Sutskever, I., Salakhutdinov, R.:
  Dropout: {{A Simple Way}} to {{Prevent Neural Networks}} from
  {{Overfitting}}. Journal of Machine Learning Research  \textbf{15}(56),
  1929--1958 (2014)

\bibitem{vaswaniAttentionAllYou2017a}
Vaswani, A., Shazeer, N., Parmar, N., Uszkoreit, J., Jones, L., Gomez, A.N.,
  Kaiser, {\L}., Polosukhin, I.: Attention is {{All}} you {{Need}}. In:
  Advances in {{Neural Information Processing Systems}}. vol.~30. {Curran
  Associates, Inc.} (2017)

\bibitem{velickovicGraphAttentionNetworks2018}
Veli{\v c}kovi{\'c}, P., Cucurull, G., Casanova, A., Romero, A., Li{\`o}, P.,
  Bengio, Y.: Graph {{Attention Networks}}. In: International {{Conference}} on
  {{Learning Representations}} (Feb 2018)

\bibitem{vu2020pgm}
Vu, M., Thai, M.T.: Pgm-explainer: Probabilistic graphical model explanations
  for graph neural networks. Advances in neural information processing systems
  \textbf{33},  12225--12235 (2020)

\bibitem{yangRevisitingSemisupervisedLearning2016}
Yang, Z., Cohen, W.W., Salakhutdinov, R.: Revisiting semi-supervised learning
  with graph embeddings. In: Proceedings of the 33rd {{International
  Conference}} on {{International Conference}} on {{Machine Learning}} -
  {{Volume}} 48. pp. 40--48. {{ICML}}'16, {JMLR.org}, {New York, NY, USA} (Jun
  2016)

\bibitem{ying2019gnnexplainer}
Ying, Z., Bourgeois, D., You, J., Zitnik, M., Leskovec, J.: Gnnexplainer:
  Generating explanations for graph neural networks. Advances in neural
  information processing systems  \textbf{32} (2019)

\bibitem{yuan2022explainability}
Yuan, H., Yu, H., Gui, S., Ji, S.: Explainability in graph neural networks: A
  taxonomic survey. IEEE Transactions on Pattern Analysis and Machine
  Intelligence  (2022)

\bibitem{zeilerVisualizingUnderstandingConvolutional2014a}
Zeiler, M.D., Fergus, R.: Visualizing and {{Understanding Convolutional
  Networks}}. In: Fleet, D., Pajdla, T., Schiele, B., Tuytelaars, T. (eds.)
  Computer {{Vision}} \textendash{} {{ECCV}} 2014. pp. 818--833. Lecture
  {{Notes}} in {{Computer Science}}, {Springer International Publishing},
  {Cham} (2014). \doi{10.1007/978-3-319-10590-1\_53}

\bibitem{zhouLearningDeepFeatures2016}
Zhou, B., Khosla, A., Lapedriza, A., Oliva, A., Torralba, A.: Learning {{Deep
  Features}} for {{Discriminative Localization}}. In: 2016 {{IEEE Conference}}
  on {{Computer Vision}} and {{Pattern Recognition}} ({{CVPR}}). pp.
  2921--2929. {IEEE Computer Society} (Jun 2016). \doi{10.1109/CVPR.2016.319}

\end{thebibliography}

% For bibLaTeX users:
%\printbibliography

\appendix

\section{Loyalty with all neighbors deleted for GNNs with self-loops}
\label{sec: loyaty with all neighbors deleted for gnns with self-loops}

In this section the loyalties when all
neighbors with non-zero importance (according to each explainability technique) are deleted are compared with the loyalties when all neighbors are deleted for GNNs with self-loops. Results can be visualized in Table~\ref{tab: loyalty with all neighbors deleted with self-loops}.

\begin{table}[h]
	\centering
	\caption{Loyalty with all neighbors deleted for GNNs with self-loops}
	\label{tab: loyalty with all neighbors deleted with self-loops}
	\scalebox{0.9}{
    	\begin{tabular}{ccccccccccccc}
    		\toprule
    		\multirow{2.5}{*}{Method} & \multicolumn{2}{c}{Cora} & \multicolumn{2}{c}{CiteSeer} & \multicolumn{2}{c}{PubMed}  \\
    		\cmidrule(lr){2-3} \cmidrule(lr){4-5} \cmidrule(lr){6-7}
    	    & GCN & GAT & GCN & GAT & GCN & GAT \\
    		\midrule
    		Saliency Map & 0.73 & 0.59 & 0.76 & 0.70 & 0.81 & 0.82 &\\
    		Smoothgrad & 0.73 & 0.59 & 0.76 & 0.70 & 0.81 & 0.82 &\\
    		Deconvnet & 0.73 & 0.59 & 0.76 & 0.70 & 0.81 & 0.82 &\\
    		Guided Backprop & 0.73 & 0.59 & 0.76 & 0.70 & 0.81 & 0.82 &\\
    		GNNExplainer & 0.73 & 0.59 & 0.76 & 0.69 & 0.81 & 0.82 &\\
    		PGExplainer & 0.73 & 0.59 & 0.76 & 0.70 & 0.81 & 0.82 &\\
    		Without Neighbors & 0.73 & 0.59 & 0.76 & 0.70 & 0.81 & 0.82 &\\
    		\bottomrule
    	\end{tabular}
    }
\end{table}

Based on the loyalty when all the neighbors have been deleted, it is clear that the loyalty only decreases to a certain point due to self-loops. That makes the models classify correctly for certain nodes based only on the node that is being classified, since all its neighbors have been deleted. It is also important to note that unlike in GNNs without self-loops (shown in Section~\ref{sec: experiments and results}) all explainability methods are able to find all important neighbors, since the loyalties when all neighbors are deleted are the same, or almost the same, as the ones when all neighbors with non-zero importance are deleted. 

\section{Not finding the neighbors without self-loops}
\label{sec: not finding neighbors}
In this appendix, an example where techniques that use the gradients for their computations are not able to discover all the important neighbors will be provided. Furthermore, the example will prove how a neighbor can contribute and still have a zero gradient. 

Specifically, the example will be a classification of an undirected graph with a two layer GNN, as the case of the citation network datasets used in this paper. Let node $i$ be the node that is being classified and node $j$ a first-hop neighbor only reachable through $i$, i.e. it is not a second-hop neighbor of $i$ through another node. The relation between the nodes is bidirectional. If there are no self-loops in a two layer GNN, the gradient of node $i$ with respect to $j$ will be zero (only gradients with respect to second-hop neighbors will be non-zero). However, node $i$ could send a message in the forward pass to $j$ that is sent back in the second GNN layer to $i$, helping the classification of node $i$. That way node $j$ would have contributed to classification and still have a zero gradient.

The example provided is just to show how a neighbor could contribute having a zero gradient. Examples like this can be constructed for directed and undirected graphs, making techniques that use the gradients in its computations fail in finding all the important neighbors when there are no self-loops.

\section{Performance on test set}
\label{sec: performance on test set}

In this Section the performance on the datasets test sets used in this paper can be visualized (Table~\ref{tab: accuracies of test data}). It is important to note that these models were not chosen because they achieve state of the art results, since there could be better models as in \cite{izadiOptimizationGraphNeural2020}, but because we find that there are commonly used architectures in GNNs.

\makeatletter
\setlength{\@fptop}{0pt}
\makeatother

\begin{table}[h]
	\centering
	\caption{Accuracies of test data}
	\label{tab: accuracies of test data}
	\scalebox{0.9}{
    	\begin{tabular}{ccccccc}
    		\toprule
    		\multirow{2.5}{*}{Self-Loops} & \multicolumn{2}{c}{Cora} & \multicolumn{2}{c}{CiteSeer} & \multicolumn{2}{c}{PubMed}  \\
    		\cmidrule(lr){2-3} \cmidrule(lr){4-5} \cmidrule(lr){6-7}
    		& GCN & GAT & GCN & GAT & GCN & GAT \\
    		\midrule
    		With & 0.81  & 0.76 & 0.64 & 0.62 & 0.78 & 0.77 \\
    		Without & 0.80  & 0.72 & 0.59 & 0.61 & 0.78 & 0.75 \\
    		\bottomrule
    	\end{tabular}
    }
\end{table}

\end{document}